%% file: tmlr.tex
\documentclass[10pt]{article} 
\usepackage[preprint]{tmlr}

\input{math_commands.tex}

\usepackage{hyperref}
\usepackage{url}
\usepackage{graphicx} 
\usepackage{subcaption}
\usepackage{booktabs}
\usepackage{multirow}

\title{The Gaussian-Multinoulli Restricted Boltzmann Machine: A Potts Model Extension of the GRBM}

\author{\name Nikhil Kapasi \email nkapasi@ucsb.edu\\
      \addr Department of Electrical and Computer Engineering\\
      University of California, Santa Barbara
      \AND
      \name Mohamed Elfouly \email elfouly@ucsb.edu\\
      \addr Department of Electrical and Computer Engineering\\
      University of California, Santa Barbara
      \AND
      \name William Whitehead \email williamwhitehead@ucsb.edu\\
      \addr Department of Electrical and Computer Engineering\\
      University of California, Santa Barbara
      \AND
      \name Luke Theogarajan \email lusthe@ucsb.edu\\
      \addr Department of Electrical and Computer Engineering\\
      University of California, Santa Barbara
      }

\def\month{MM}  
\def\year{YYYY} 
\def\openreview{\url{}} 

\begin{document}

\maketitle

\begin{abstract}
Many real-world tasks, from associative memory to symbolic reasoning, benefit from discrete, structured representations that standard continuous latent models can struggle to express. We introduce the Gaussian–Multinoulli Restricted Boltzmann Machine (GM-RBM), a generative energy-based model that extends the Gaussian–Bernoulli RBM (GB-RBM) by replacing binary hidden units with $q$-state categorical (Potts) units, yielding a richer latent state space for multivalued concepts. We provide a self-contained derivation of the energy, conditional distributions, and learning rules, and detail practical training choices (contrastive divergence with temperature annealing and intra-slot diversity constraints) that avoid state collapse. To separate architectural effects from sheer latent capacity, we evaluate under both capacity-matched and parameter-matched setups, comparing GM-RBM with GB-RBM configured to have the same number of possible latent assignments. On analogical recall and structured memory benchmarks, GM-RBM achieves competitive, and in several regimes, improved recall at equal capacity with comparable training cost, despite using only Gibbs updates. The discrete $q$-ary formulation is also amenable to efficient implementation. These results clarify when categorical hidden units provide a simple, scalable alternative to binary latents for discrete inference within tractable RBMs.
\end{abstract}

\section{Introduction}
Restricted Boltzmann Machines (RBMs) are energy-based models with an undirected bipartite graph structure with visible and hidden units and a subset of the broader fully connected Boltzmann machines. The lack of intralayer connections enables training tractability through parallel updates of all units within each layer (Block Gibbs updates) through a biased stochastic gradient estimator for log-likelihood (Contrastive Divergence) \citep{hinton2002training, hinton2006reducing}. Although powerful, the strictly binary units in hidden and visible units make them non-ideal for dealing with multivalued data. One approach to extend their use to continuous data is to replace the visible units with Gaussian units, termed Gaussian--Bernoulli RBMs (GB-RBMs) \citep{liao2022gaussianbernoullirbmstears}. However, their binary hidden units often struggle with inherently categorical, mutually exclusive factors. We address this mismatch by replacing each binary hidden unit with a one-of-$q$ categorical unit (Potts), yielding the Gaussian--Multinoulli RBM (GM-RBM): Gaussian visibles paired with $q$-state latent hidden units. Conceptually, this keeps the RBM's simplicity but aligns inductive bias with categorical structure, effectively allowing the model to capture inherent, underlying categorical relationships. Prior work tended to retain binary latents as they fit existing tooling and samplers; in contrast, categorical slots raise practical issues (e.g., intra-slot degeneracy) and make fair comparisons challenging. In order to separate architectural effects from raw capacity, we design comparison protocols and clarify the challenges associated with them. 

Our contributions are threefold: (i) a drop-in Potts hidden layer that preserves tractable conditionals while retaining the standard RBM training pipeline; (ii) comparison protocols (capacity-matched and parameter-matched) that isolate the effect of categorical slots; and (iii) empirical results showing that, with equal negative-phase budgets and pure block Gibbs updates, increasing $q$ consistently improves image quality (FID) and hetero-associative recall, closing much of the gap to GB-RBMs that require costlier Gibbs–Langevin steps \citep{liao2022gaussianbernoullirbmstears}.
The key message being \emph{a minimal architectural change yields disproportionate gains} due to the q-slot augmented latent: sharper posteriors, more interpretable codes, and stronger retrieval and generation while utilizing similar compute resources. 

\section{Background and Motivation}
\subsection{Background}
Boltzmann machines (BM) are a class of energy-based models utilizing binary units ($\{-1,1\}$) that sample from a Boltzmann probability distribution over $x=(v,h)$ given by: 
\[
p_\theta(x)=\frac{1}{Z(\theta)}\exp\!\big(-E_\theta(x)\big), \qquad 
Z(\theta)=\sum_x \exp\!\big(-E_\theta(x)\big).
\]
where $\theta$ is a set of biases ($a_i$) and weights ($W_{ij}$). \citep{smolensky1986information, hinton2002training} The energy function for a BM is given by: 
\[
E_\theta(x)= -\sum_i a_i x_i - \tfrac12\sum_{i\neq j} W_{ij} x_i x_j,\qquad W_{ij}=W_{ji},\;W_{ii}=0.
\]
Log-likelihood learning decomposes into a positive (data) and negative (model) phase:
\[
\frac{\partial}{\partial\theta}\log p_\theta(v)
= \mathbb{E}_{p(h\mid v)}\!\Big[-\tfrac{\partial E(v,h)}{\partial\theta}\Big]
- \mathbb{E}_{p(v\mid h)}\!\Big[-\tfrac{\partial E(v,h)}{\partial\theta}\Big],
\]
which yields the classic correlation-matching rules: 
\[
\frac{\partial}{\partial W_{ij}}\log p_\theta(v)=\langle x_i x_j\rangle_{\text{data}}-\langle x_i x_j\rangle_{\text{model}},
\qquad
\frac{\partial}{\partial a_i}\log p_\theta(v)=\langle x_i\rangle_{\text{data}}-\langle x_i\rangle_{\text{model}}.
\]

However, correlation-matching rules are intractable for most real-world data \citep{liao2022gaussianbernoullirbmstears}. One way to overcome this is to use the Restricted Boltzmann machine: 
Restricted Boltzmann machines split the graph into a bipartite structure with visible units $v$ and hidden units $h$, removing within-layer edges so that the joint probability distribution can be broken down into simpler conditional probabilities enabling block Gibbs sampling. For binary visibles $v\in\{0,1\}^n$ and binary hiddens $h\in\{0,1\}^m$,
\[
E(v,h) = -b^\top v - c^\top h - v^\top W h,
\]
\[
p(h_j=\textbf{1}\mid v)=\sigma\big((W^\top v)_j + c_j\big),\qquad
p(v_i=\textbf{1}\mid h)=\sigma\big((W h)_i + b_i\big),
\]
and the learning updates specialize to moment matching between the two layers,
\[
\Delta W \propto \mathbb{E}[v h^\top]_{\text{data}}-\mathbb{E}[v h^\top]_{\text{model}},\quad
\Delta b \propto \mathbb{E}[v]_{\text{data}}-\mathbb{E}[v]_{\text{model}},\quad
\Delta c \propto \mathbb{E}[h]_{\text{data}}-\mathbb{E}[h]_{\text{model}}.
\]
Contrastive divergence(CD) approximates model expectations by alternating $h\!\sim\!p(h\mid v)$ and $v\!\sim\!p(v\mid h)$ for a $k$ steps of data CD-$k$ or for a persistence chain (persistent CD). Usually, both hidden and visible units are taken as binary, but for continuous data it is natural to use Gaussian visible units \citep{liao2022gaussianbernoullirbmstears}. In the Gaussian–Bernoulli RBM (Hinton), visibles are Gaussian with diagonal variance $\sigma^2$ and hiddens remain binary. A convenient parameterization is:
\[
E(v,h)= \sum_{i}\frac{(v_i-\mu_i)^2}{2\sigma_i^2}
- \sum_{i,j}\frac{v_i}{\sigma_i^2} W_{ij} h_j
- \sum_j b_j h_j,
\]
which gives closed-form conditionals used inside CD,
\[
p(v\mid h)=\mathcal{N}\!\big(\mu + W h,\ \mathrm{diag}(\sigma^2)\big),\qquad
p(h_j=1\mid v)=\text{Sigmoid}\!\Big(\big[W^\top (v\oslash \sigma^2)\big]_j + b_j\Big).
\]
The corresponding stochastic-gradient updates keep the same positive/negative structure but respect the visible scaling,
\[
\Delta \boldsymbol{W} \propto \Big\langle \left( \frac{\mathbf{v}}{\boldsymbol{\sigma}^2} \right)\, \mathbf{h}^\top \Big\rangle_{\text{data}}
- \Big\langle \left( \frac{\mathbf{v}}{\boldsymbol{\sigma}^2} \right)\, \mathbf{h}^\top \Big\rangle_{\text{model}},
\]
\[
\Delta \boldsymbol{\mu} \propto \Big\langle \frac{\mathbf{v}-\boldsymbol{\mu}}{\boldsymbol{\sigma}^2} \Big\rangle_{\text{data}}
- \Big\langle \frac{\mathbf{v}-\boldsymbol{\mu}}{\boldsymbol{\sigma}^2} \Big\rangle_{\text{model}}
\]

\[
\Delta \boldsymbol{b} \propto \langle \boldsymbol{h}\rangle_{\text{data}}-\langle \boldsymbol{h}\rangle_{\text{model}}
\qquad
\]
In this work we extend the role of the binary hidden variables by replacing each Bernoulli with a one-of-$q$ categorical (Potts) slot, aligning the latent prior with mutually exclusive factors while preserving the RBM’s block-sampling tractability and the Gaussian visible layer.

\subsection{Motivating Principles}
Many perceptual and symbolic factors are naturally \emph{categorical and mutually exclusive}. Approximating such structure with many Bernoulli latents (as in a GB–RBM) forces variants to be represented by co–activating subsets of units, which encodes information across the hidden layer and yields ambiguous codes.

The Gaussian–Multinoulli RBM (GM–RBM) encodes each factor as a \emph{one-of-$q$} slot. With $m$ slots and hidden configuration $h=(h_1,\dots,h_m)$ where $h_j\in\{1,\dots,q\}$, the visible conditional remains Gaussian,
\[
p(v\mid h)=\mathcal{N}\!\Big(b + \sum_{j=1}^{m} W^{(h_j)}_{:,j},\, I\Big),
\]
so each chosen state contributes a single template vector and the mean is a sum of selected templates. This preserves the RBM’s locally linear structure encoding the continuous variables using the set of latent discrete variables. A key advantage of the Potts node is that each slot has its own weight vector, so swapping slots does not alter the previous slot's learned weight. In practice, this yields sharper posteriors and more degrees of freedom than simply enforcing a one-hot encoding across multiple Bernoullis, which intrinsically lack intra-layer coupling.

GM–RBM is a modification where we replace binary hidden units with categorical slots but retain Gaussian visibles and standard RBM conditionals. We assess GM–RBM on hetero-associative recall and image modeling. To highlight the intrinsic strength of the GMRBM to the GBRBM, we develop comparison protocols that are \emph{capacity-matched (hidden nodes)} and \emph{parameter-matched (weights)}, which appear later. 

\section{Theoretical Foundations of the Gaussian--Multinoulli RBM (GM-RBM)}

The Gaussian--Multinoulli RBM (GM-RBM) replaces binary hidden units with discrete $q$-state categorical variables while keeping a continuous visible layer. This yields a combinatorial latent space with simple, closed-form conditionals.

\subsection{Notation}
Let $v\in\mathbb{R}^{n}$ be the visible vector. The hidden code is $h=(h_1,\dots,h_m)$ with $h_j\in\{1,\dots,q\}$. Parameters are: visible bias $b\in\mathbb{R}^{n}$; hidden bias $c_{j,k}\in\mathbb{R}^{m}$; and state-specific templates $W^{(k)}_{:,j}\in\mathbb{R}^{n}$. Define the conditional mean

$$
\mu(h) = b + \sum_{j=1}^{m} W^{(h_j)}_{:,j}
$$

\subsection{Energy, Joint, and Conditionals}
The energy is

$$
E(v,h) = \tfrac{1}{2}\sum_{i=1}^{n} (v_i - b_i)^2 - \sum_{j=1}^{m} c_{j,h_j} - \sum_{i=1}^{n}\sum_{j=1}^{m} W^{(h_j)}_{ij}\, v_i.
$$

Completing the square gives $E(v,h)=\tfrac{1}{2}\lVert v-\mu(h)\rVert_2^2 + K(h)$ with $K(h)=\tfrac{1}{2}(\lVert b\rVert_2^2-\lVert \mu(h)\rVert_2^2)-\sum_{j} c_{j,h_j}$. The joint is $p(v,h)\propto\exp(-E(v,h))$. The conditionals are

$$
 p(v\mid h)=\mathcal{N}\big(\mu(h), \boldsymbol{1}\big)
$$
$$
 \qquad p(h_j=k\mid v)=\frac{\exp\big(c_{j,k}+(W^{(k)}_{:,j})^{\top}v\big)}{\sum_{k'=1}^{q}\exp\big(c_{j,k'}+(W^{(k')}_{:,j})^{\top}v\big)} =\text{Softmax}{(c_{j,k'}+(W^{(k')}_{:,j})^{\top}v\big)}
$$

\subsection{Architecture and Special Cases}
Each slot contributes one of $q$ templates, so the codebook $\{\mu(h): h\in\{1,\dots,q\}^{m}\}$ has size $q^{m}$. When $q=2$ and parameters are tied as $W^{(1)}_{:,j}-W^{(2)}_{:,j}=\widetilde W_{:,j}$ and $c_{j,1}-c_{j,2}=\widetilde b_j$ with a corresponding recentering of $b$, the GM-RBM reduces to a Gaussian--Bernoulli RBM. There is no requirement that $q$ be even; experiments may vary $q$ according to task. It is also possible to use usual variance reweighting as opposed to unit variance used in the GB-RBM.

\subsection{Parameter Count and Fair Comparison Protocols}
GM-RBM has $n$ parameters for $b$, plus $m q$ for $c$, plus $n m q$ for $W$, for a total of $n + m q\,(1+n)$. A GB-RBM with $m'$ binaries has $n + m' + n m'$. We evaluate under two protocols:
\begin{itemize}
\item \textit{Parameter-matched}: Match the number of latent assignments by setting $m' \approx m\,\log_{2} q$, since $|\mathcal{H}_{\text{GM}}| = q^{m}$ and $|\mathcal{H}_{\text{GB}}| = 2^{m'}$. Essentially, choose $m'$ so total parameters are comparable across models
\item \textit{Capacity-matched}: Choose $|\mathcal{H}_{\text{GB}}|$ in such a way that it matches the total number of learned representation for a given $|\mathcal{H}_{\text{GM}}|$
\end{itemize}
These regimes separate architectural effects (slot exclusivity and templating) from raw codebook size.

\subsection{Learning and Negative-phase Sampling}
The log-likelihood gradient satisfies

$$
 \tfrac{\partial}{\partial \theta}\log p(v)=\mathbb{E}_{p(h\mid v)}\Big[-\tfrac{\partial E(v,h)}{\partial \theta}\Big]-\mathbb{E}_{p(v\mid h)}\Big[-\tfrac{\partial E(v,h)}{\partial \theta}\Big],\quad \theta\in\{b,c,W\}.
$$

We approximate the model expectation with short Markov chains.

\paragraph{Block Gibbs.} Alternate $h\sim p(h\mid v)$ using the per-slot softmax and $v\sim p(v\mid h)=\mathcal{N}(\mu(h), \mathbf{1})$. The visible draw is exact and parameter free.

\paragraph{Gibbs with visible Langevin.} 
Some implementations replace the exact Gaussian draw with an unadjusted Langevin step using $\nabla_v\log p(v\mid h)=\mu(h)-v$ \citep{liao2022gaussianbernoullirbmstears}:

$$
 v_{t+1}=v_t+\tfrac{\varepsilon^2}{2}\big(\mu(h_t)-v_t\big)+\varepsilon\,\xi_t,\quad \xi_t\sim\mathcal{N}(0,\mathbf{1}),\ \varepsilon>0.
$$

This introduces a stepsize–dependent discretization error; for fixed $h_t$, multiple Langevin steps only approximate the exact conditional. In many Gaussian RBM variants, each visible dimension is allowed its own variance, so that $p(v\mid h)=\mathcal{N}(\mu(h),\Sigma)$ with a diagonal covariance $\Sigma = \operatorname{diag}(\sigma^2)$. In this case, the gradient becomes $\Sigma^{-1}(\mu(h_t)-v_t)$, and the drift term is modified to $(\mu(h_t) - v_t)/\sigma^2$ (applied elementwise). Intuitively, inserting a visible Langevin move between hidden-layer updates allows hidden units to exchange information indirectly through the visible layer, encouraging more cooperative encodings and improving the model’s ability to capture complex dependencies.

 In contrast to the expensive Langevin update, we use exact block Gibbs updates for the GM-RBM. The reason we utilize this less expensive block Gibbs update is that the visible Langevin step is merely an approximate sampler for the same conditional distribution $p(v\mid h)$. It does not inject additional information into the model, but instead introduces stepsize-dependent bias at an extra computational cost. Meanwhile, the multi-state multinoulli latent variables already enable this information to be represented and shared across hidden units through standard block Gibbs updates, making an additional visible Langevin refinement unnecessary in our setting. To ensure fair comparison to GB-RBM baselines, we explicitly report whether a visible Langevin step is used, the stepsize, the number of steps, and whether chains are persistent. 

\paragraph{Sampler cost, mixing, and our choice.} The visible Langevin variant adds at least one extra update and a stepsize hyperparameter per negative step, which increases compute relative to an exact Gaussian draw. In our setting the Gibbs visible update samples exactly from $N(\mu(h),\mathbf{1})$ with a single noise vector and no stepsize tuning. The GM-RBM did was able to achieve quality results without resorting to the more expensive Langevin update. This choice also highlights a modeling claim of this work: with slot-wise categorical latents, GM-RBM exhibits fast mixing under basic Gibbs updates without relying on heavier samplers. We report chains and effective sample diagnostics to substantiate this point.

\subsection{Key Properties}
\begin{itemize}
\item Locally linear: given $h$, $p(v\mid h)$ is Gaussian with fixed covariance.
\item Globally discrete: the means form a finite codebook indexed by discrete slots.
\item Modular: slots contribute additively in $\mu(h)$ and independently in $p(h\mid v)$.
\end{itemize}

\section{Hetero-associative Memory}
Hetero-associative memory refers to a system’s capability to learn paired associations between distinct patterns, allowing the retrieval of a target pattern (e.g., a response word) when presented with a corresponding stimulus (e.g., a cue word) \citep{BiAM, EAM}. This concept, rooted in cognitive modeling and neural computation, was further explored by Hinton in 1981 \citet{parAM} and later employed in language-related tasks using Gaussian-Bernoulli Restricted Boltzmann Machines (GB-RBMs) \citep{tsutsui2019analog}. However, the binary hidden units of GB-RBMs impose a hard ceiling on representational capacity.

Across the board, GM-RBMs deliver consistently higher recall than GB-RBMs, with gains amplifying at larger $q$ and corpus sizes even when the number of parameters were identical between the two models. It is important to note that this performance gain is obtained although the GM-RBM was trained using only \emph{Gibbs sampling}, while the GB-RBM used a stronger and more expensive hybrid Gibbs-Langevin Sampling.

\subsection{Experimental Setup}
We replicated the experimental setup in Tsutsui and Hagiwara, constructing a word-pair dataset representing conceptual relationships (e.g., "apple is-a fruit") \citep{tsutsui2019analog}. We randomly selected pairs from WordNet (\citet{wordnet}), excluding compound or incomplete entries and sampled 500-3,000 word pairs to create smaller datasets for training and testing scalability.

Each word pair is treated as a directional association task and encoded as a concatenated embedding vector \citep{mikolov2013efficient}, mirroring the hetero-associative memory objective of recalling a target concept from a given stimulus. 

We trained a 200-dimensional Continuous Bag of Words (CBOW) Word2Vec model \citep{mikolov2013efficient} on the word-pair dataset (100 iterations, window size 5, no frequency cutoff). We normalized each vector dimension to zero mean and unit variance to mitigate small-magnitude issues. For each stimulus-response pair, embeddings were concatenated to form a 400-dimensional input vector for the RBM's visible layer.

We used a compute node (Intel Xeon Gold 6154 ×2, 512 GB RAM, NVIDIA Tesla P40) running Red Hat Enterprise Linux 8. Models were trained with CUDA-accelerated PyTorch (v1.13) on a single GPU. Experiments were automated via a modular framework for data loading, configuration, visualization, and checkpointing.

\subsection{Training}
Our GM-RBM learned to associate stimulus–response word embeddings. Datasets comprised semantically related word pairs (e.g., doctor–nurse, sun–light). Following a similar procedure to Tsutsui and Hagiwara \citet{tsutsui2019analog}, we trained a CBOW Word2Vec model \citep{rehurek_lrec} on these two-word sentences to capture in-domain semantics, producing 200-dimensional embeddings (100 iterations, window size 5, no frequency cutoff).

We normalized each embedding dimension to zero mean and unit variance to avoid numerical instability. The stimulus and response embeddings were concatenated into a 400-dimensional visible layer vector.

We trained the GB-RBM using contrastive divergence with a two-step Gibbs burn-in, employing the more expensive Gibbs–Langevin sampling procedure. In contrast, the GM-RBM variant relied solely on standard Gibbs sampling, reducing computational overhead. Notably, in the special case of $q=2$, the GM-RBM formulation becomes almost equivalent to the GB-RBM: the only difference is that in the GB-RBM, the two weight matrices are essentially negatives of one another. Hidden-unit counts were scaled inversely with the number of Potts states to keep capacity constant. Training used Adam (LR = $10^{-4}$) with mini-batches of 64. We evaluated recall by inferring responses via Gibbs sampling and selecting the nearest neighbor; accuracy was the percentage of correct matches.

We stopped training early when recall accuracy reached 0.98 on the validation set, when the standard deviation of validation accuracy over 20 checkpoints fell below 0.01, or when no improvement was observed for 10 consecutive checkpoints.

\subsection{Results}
 To isolate the effect of the Potts hidden units, we did two key tests. The first was a parameter-matched comparison where the total number of weights was kept equal while $q$ was increased. This ensured that any observed gains in recall accuracy arise purely from the richer Potts representation. The second was a similar sweep where $q$ was held constant while the number of hidden units themselves was increased. Both sweeps were done across varying dataset sizes, measuring the accuracy of the recall in pairs embedded in Word2Vec \citep{mikolov2013efficient}.

Each marker in Figures \ref{fig:RetAccPottsStates} and \ref{fig:hiddensweep} denotes an independently trained model using identical early-stopping criteria. Error bars have been omitted as the observed performance gains were uniformly large, making statistical uncertainty negligible for the comparative trends shown. 

For data shown in Figure \ref{fig:RetAccPottsStates}, it is important to note that the total number of parameters was held constant i.e. hidden layers were decreased proportionally to the cardinality of the Potts state $q$ (See \ref{tab:q_hidden_units_horizontal}). 

It is also important to note $q=2$ for the GM-RBM is a case of the Potts nodes where weights do not have to be constrained i.e. negatives of one another. The original GB-RBM outperforms the GM-RBM for $q=2$, possibly due to the lack of this constraint. Despite the GM-RBMs being trained using the simpler Gibbs update, the $q=4,6,8,10$ GM-RBM models all vastly outperform the GB-RBM, which used the more expensive Gibbs-Langevin update \citep{liao2022gaussianbernoullirbmstears}.  

Since constraining the total number of parameters to be identical might obfuscate the potential of both models, we swept the number of hidden nodes and dataset sizes as shown in Figure \ref{fig:hiddensweep}. It can be clearly seen that GM-RBM $q=2$ and the GB-RBM both fail when $N>2000$, while the GM-RBM $q=4$ maintains its retrieval accuracy. 

\subsubsection{Parameter matched \textit{q} sweep}
\begin{figure}[ht!]
  \centering
  \includegraphics[width=0.8\textwidth]{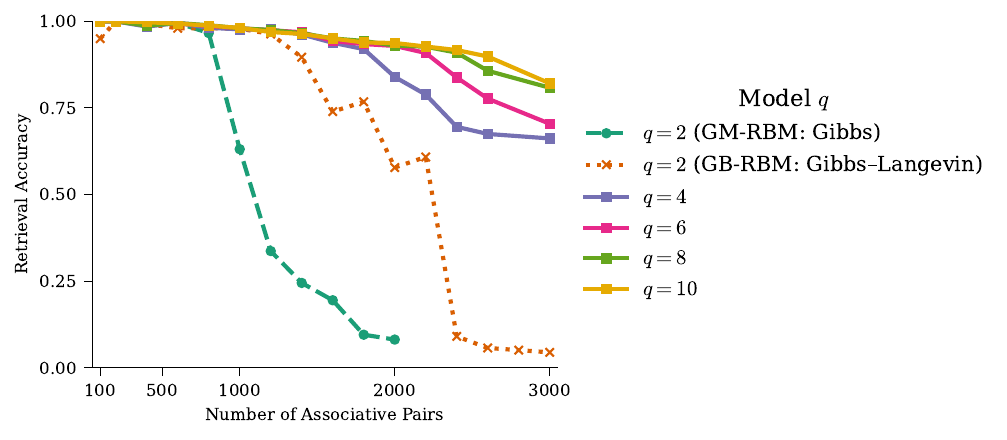}
  \caption{ Retrieval accuracy versus the number of associative pairs for different numbers of Potts states (q) in a parameter-matched GB-RBM and GM-RBM setup}
  \label{fig:RetAccPottsStates}
\end{figure}
To empirically showcase the direct effect of the additional Potts states, we held the total number of model parameters constant while varying the number of Potts states $q$, isolating the effect of the state space's structure on hetero-associative performance. Figure \ref{fig:RetAccPottsStates} plots retrieval accuracy against the number of associative pairs in the training set for each $q$. For the binary case $q=2$, the GB-RBM with Gibbs–Langevin sampling maintains near-perfect accuracy at small dataset sizes but collapses sharply beyond 1000 pairs, whereas the GM-RBM using only Gibbs sampling degrades more rapidly. In contrast, models with higher state cardinality ($q=4,6,8,10$) sustain almost perfect retrieval up to roughly 1200–1500 pairs and exhibit a more gradual decline as a function of the dataset size($N$). Notably, $q=10$ consistently outperforms lower-$q$ configurations even for large $N$. 

To keep the model capacity fixed across different Potts‐state configurations, we solve for the number of hidden units $n_h$ by dividing the total budget of weight parameters $n_w$ by the size of each hidden Potts spin $q$. Since each hidden unit with \(q\) Potts states contributes \(q\) weight vectors of length equal to the number of visible units $n_v$, then number of hidden units is given by
\begin{equation}
n_h \texttt =  \lceil \frac{n_w}{n_v}\times q\ \rceil
\end{equation}

We chose \(\mathrm{n_w} = 800{,}000\), ensuring that the binary case $q=2$ matched the optimal hidden‐unit count of 1,000 reported by Tsutsui and Hagiwara \citet{tsutsui2019analog}for the GB-RBM. This ensures that, as we increase the number of Potts states \(k\), we reduce the number of hidden units proportionally so that every model has the same total number of parameters.

\begin{table}[ht]
\caption{Potts-state \(q\) versus number of hidden units}
\label{tab:q_hidden_units_horizontal}
\begin{center}
\begin{tabular}{lccccc}
\multicolumn{1}{c}{} &
\multicolumn{1}{c}{\bf $q=2$} &
\multicolumn{1}{c}{\bf $q=4$} &
\multicolumn{1}{c}{\bf $q=6$} &
\multicolumn{1}{c}{\bf $q=8$} &
\multicolumn{1}{c}{\bf $q=10$}
\\ \hline \\
Hidden Units & 1000 & 500 & 333 & 250 & 200 \\
\end{tabular}
\end{center}
\end{table}

This result demonstrates increasing the number of discrete states enhances memory robustness under constant parameter constraints. By allocating representational capacity across more Potts states, the model tolerates larger associative loads, suggesting an optimal trade-off point around moderate state sizes (e.g., $q=6$ or $8$) for typical dataset scales in the 2000–2500 pair range.

\subsubsection{Hidden nodes sweep}
\begin{figure}[ht!]
  \centering
  \includegraphics[width=0.95\textwidth]{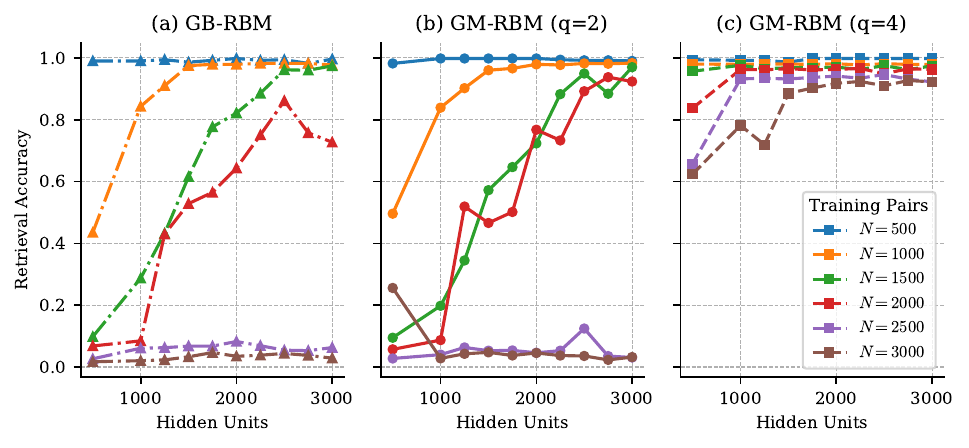}
  \caption{Retrieval accuracy on a semantic hetero‐associative memory task as a function of hidden‐layer size for different model variants and dataset sizes. (a) GB-RBM (Gibbs--Langevin Update), (b) GM-RBM with $q=2$ (Gibbs Update), and (c) GM-RBM with $q=4$ (Gibbs Update). Each curve corresponds to a different number of training word‐pair examples ($N$).}
  \label{fig:hiddensweep}
\end{figure}
In order to showcase how varying the hidden-layer dimensionality affects retrieval performance across dataset sizes, we swept the number of hidden units from 500 to 3000 while fixing the state cardinality $q$. Figure \ref{fig:hiddensweep} presents retrieval accuracy curves for three models: a standard Gaussian–Bernoulli RBM (GB-RBM), a GM-RBM with $q=2$, and a GM-RBM with $q=4$). Each curve represents a different dataset size ($N=500,1000,1500,2000,2500,3000$ associative word pairs).

For the binary GM-RBM ($q=2$), accuracy is near-perfect at small loads but degrades sharply as $N$ increases, recovering only when hidden units exceed 1500. In contrast, the Potts-based GM-RBM with $q=4$ maintains over 90\% accuracy across all dataset sizes with just 1000 hidden units. The GB-RBM baseline requires roughly 2500 hidden units to achieve similar performance at large scales (e.g., $N=2500$).

These findings demonstrate a clear trade-off between hidden-layer dimensionality and state complexity: increasing $q$ substantially lowers the hidden-unit requirement for robust associative recall, yielding a more parameter-efficient memory architecture.

\section{Auto-associative memory}
As a "proof-of-concept" to demonstrate the generative capability of our GM-RBM, we performed a replication of the generative experiments from the original GB-RBM paper \citet{liao2022gaussianbernoullirbmstears}. On MNIST, we trained for 500 epochs, and on CelebA for 100 epochs. Starting from i.i.d.\ Gaussian noise in the visible layer and running 1,000 steps of Gibbs sampling, we obtained samples (see Figure~\ref{fig:sampledmnistCelebA}) that, by visual inspection, showcasing the GM-RBM's ability to still produce high-quality generative outputs.

\subsection{Experimental setup}
We target two key datasets to showcase the GM-RBM's generative abilities: \textbf{MNIST} (28×28 grayscale handwritten digits) \citep{lecun1998mnist} and \textbf{CelebA} (Center-cropped and resized RGB facial images at 64×64 resolution) ~\citep{liu2015deep}.

All image datasets are normalized to zero mean and unit variance per channel~\citep{liao2022gaussianbernoullirbmstears}. CelebA images are center-cropped to 140×140 before downsampling to 64×64~\citep{liu2015deep}. The GMM datasets are constructed using predefined priors and covariance structures, allowing for exact density visualization and sampling~\citep{liao2022gaussianbernoullirbmstears}. For each dataset, we generate training and evaluation splits using standard protocols.

Experiments were performed on a dedicated compute node equipped with dual Intel Xeon Gold 5218 CPUs, 512 GB of RAM, and eight NVIDIA RTX 6000 GPUs connected via NVLink. All models were trained using CUDA-accelerated PyTorch (v1.13+) with single-GPU execution unless otherwise specified.

\subsection{Results}
\begin{figure}[h!]
\centering
\begin{subfigure}[t]{0.28\linewidth}
    \centering
    \includegraphics[width=\linewidth]{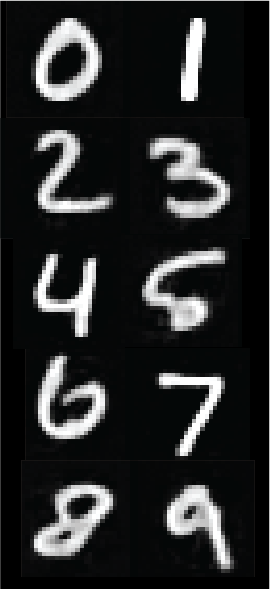}
    \caption{Selected samples from MNIST}
    \label{fig:mnist}
\end{subfigure}
\hfill
\begin{subfigure}[t]{0.62\linewidth}
    \centering
    \includegraphics[width=\linewidth]{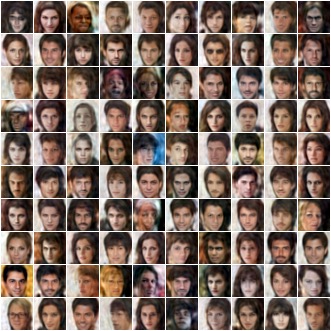}
    \caption{Generation of CelebA images from random noise with $q=4$}
    \label{fig:celebA}
\end{subfigure}

\caption{Sampled results from GM-RBM of MNIST and CelebA datasets.}
\label{fig:sampledmnistCelebA}
\end{figure}

We observe that the $q=4$ GM-RBM begins to generate visually identifiable face/digit samples (See Figure \ref{fig:mnist} and \ref{fig:celebA}) with an order of magnitude lower number of epochs compared to the GB-RBM trained with Gibbs--Langevin sampling (See Table \ref{tab:hyperparams_both}. Although a fully controlled head-to-head comparison under identical hyperparameter budgets (see Table~\ref{tab:hyperparams_both}) requires further investigation (See Section \ref{Limits}), it is important to emphasize two key distinctions. Our GM-RBM relies solely on conventional Gibbs sampling, whereas Hinton's GB-RBM employs the more computationally expensive Gibbs--Langevin procedure \citep{liao2022gaussianbernoullirbmstears}. This ability is most likely due to the rapid mixing of Potts models compared to binary Boltzmann machines \citep{potts1952, Whitehead_2023, Wureview1982.54.235}. In addition, we maintain the same total number of parameters while increasing the number of Potts states.

\begin{table}[h!]
\caption{Hyperparameter settings for GB-RBM and GM-RBM on MNIST and CelebA}
\label{tab:hyperparams_both}
\begin{center}
\begin{tabular}{lcccc}
\multicolumn{1}{c}{}                &
\multicolumn{2}{c}{\bf MNIST}       &
\multicolumn{2}{c}{\bf CelebA}      \\
\multicolumn{1}{c}{\bf Hyperparameter} &
\multicolumn{1}{c}{\bf GB-RBM}      &
\multicolumn{1}{c}{\bf GM-RBM}      &
\multicolumn{1}{c}{\bf GB-RBM}      &
\multicolumn{1}{c}{\bf GM-RBM}
\\ \hline \\
Number of states    & 2               & 4               & 2               & 4               \\
Sampling style      & Gibbs--Langevin & Gibbs           & Gibbs--Langevin & Gibbs           \\
Hidden nodes        & 4096            & 2048            & 10000           & 5000            \\
Visible nodes       & 784             & 784             & 3072            & 3072            \\
Epochs trained      & 3000            & 500             & 10000           & 100             \\
\end{tabular}
\end{center}
\end{table}

This reduction in both sampling complexity and overall training budget underscores the richer latent representational capacity afforded by Potts-style hidden units, which can capture more nuanced multimodal structure with fewer resources. The stark difference in hyperparameters was chosen specifically to highlight the GM-RBM's ability to achieve high-quality samples under a dramatically reduced resource budget, showcasing its practical efficiency.

\subsubsection{Quantitative sample quality (FID)}
\begin{table}[h]
\caption{FID (↓) under parameter-matched budgets. Best overall in \textbf{bold}.}
\label{tab:fid_capacity}
\begin{center}
\begin{tabular}{lcc}
\multicolumn{1}{c}{\bf Model} &
\multicolumn{1}{c}{\bf Potts States} &
\multicolumn{1}{c}{\bf FID}
\\ \hline \\
GM--RBM & $q=2$ & 67.08 \\
GM--RBM & $q=4$ & 56.09 \\
GM--RBM & $q=6$ & \textbf{53.07} \\
GB--RBM & N/A   & 60.06 \\
\end{tabular}
\end{center}
\end{table}
To complement the qualitative samples, we report Fréchet Inception Distance \citep{heusel2018ganstrainedtimescaleupdate} (FID; ↓ is better) under a \emph{capacity-matched} protocol ($m' \approx m \log_2 q$), holding negative-phase budgets fixed across models. Results show that increasing $q$ consistently improves GM–RBM sample quality under a pure Gibbs sampler, narrowing the gap to GB–RBM with Gibbs–Langevin despite the latter’s higher per-step compute.

With the Gibbs-only updates, GM--RBM ($q{=}6$) outperforms GB--RBM by $\approx$7 FID points (53.07 vs 60.06). With a more expensive update, it is possible to lower the FID score. It is important to note that the CelebA dataset does not necessarily have clear relational qualities, meaning that these scores are really showcasing the mixing ability of the different sampling algorithms and models. 

\section{Limitations and future work} \label{LFW}
While the Gaussian–Multinoulli RBM (GM–RBM) demonstrates stronger discrete expressiveness and robust recall under fair matching, important limitations and next steps remain.

\subsection{Limitations of the current model} \label{Limits}
\textbf{1. Sampling constraints and rationale.}
Our GM–RBM uses \emph{pure block Gibbs} (exact Gaussian visible draw + per-slot softmax posteriors). We intentionally avoid visible-space Langevin during training because it adds step-size hyperparameters, extra updates, and discretization error, increasing compute relative to a single exact Gaussian draw. 

\textbf{2. Evaluation scope and scaling.}
Most of our positive results are on hetero-associative recall with in-domain Word2Vec embeddings and proof-of-concept image generation. This leaves open: (a) robustness to stronger or off-the-shelf embeddings (e.g., large CBOW/skip-gram trained on broader corpora), (b) transfer to other modalities (audio, trajectories, time series), and (c) deeper stacks (e.g., GM front-ends for DBMs/DBNs). We also observe retrieval degradation as $N$ grows in some settings; diagnosing whether this arises from chain mixing, capacity-matching choices, or dataset effects will require ablations (burn-in length, persistence, $q$ vs.\ $m$, negative-phase steps).

\textbf{3. Training budget and stability.}
Our generative experiments used short schedules (e.g., 500 epochs on MNIST, 100 on CelebA) to highlight practicality under a reduced budget. Longer training with early-stopping and variance monitoring could reveal stability regimes, impacts on sample diversity, and when heavier negative-phase samplers pay off.

\subsection{Broader impact}
\subsubsection{Potts units in energy transformers}
Energy Transformers minimize a global energy via recurrent updates. Replacing binary hidden units with $q$-state Potts slots increases latent capacity from $2^H$ to $q^H$ and reduces attractor overlap:
\begin{equation}
    E_{\mathrm{mem}}(h) \;=\; -\sum_{\mu=1}^M\sum_{i=1}^H \delta(h_i,\xi_i^\mu).
\end{equation}
We hypothesize improved selectivity/robustness under higher $q$ and are empirically exploring GM–RBM components as ET memories; careful comparisons will match either capacity or parameter budgets as above. \citep{schroder2024energy, hoover2023energy}

As a front-end to DBMs, a Gaussian–Potts encoder maps $\mathbb{R}^D$ into a $q^H$ latent space while preserving layer-wise Gibbs training. Future work should examine stability under deeper stacks and quantify benefits over increased Bernoulli width at matched capacity. \citep{EAM, oh2022recabvaegumbelsoftmaxvariationalinference}

Many foundational generative families, RBMs, DBNs/DBMs, Hopfield-style memories, VAEs with Bernoulli latents, Gumbel-Softmax relaxations, autoregressive binarized pixel models, BNNs, SNNs, and discrete diffusion—lean on binary sampling. Swapping binary units for Potts slots increases representational granularity and sharpens attractor basins, which \emph{can} reduce interference among stored patterns at equal capacity. A broad, capacity- or parameter-matched survey across these families is an open, high-impact direction \citep{BiAM}.

Binary and Potts one-hot codes map naturally to LUTs and bitwise logic. Sparse/event-driven readout and lightweight on-chip softmax enable efficient FPGA/ASIC/neuromorphic realizations; exploring SPAD-assisted annealing or mixed-signal implementations for categorical slots is promising future work \citep{Whitehead_2023}.

\newpage
\bibliography{tmlr}
\bibliographystyle{tmlr}

\end{document}

%% file: math_commands.tex
\usepackage{amsmath,amsfonts,bm}

\def\eqref#1{equation~\ref{#1}}

\def\1{\bm{1}}

\DeclareMathAlphabet{\mathsfit}{\encodingdefault}{\sfdefault}{m}{sl}
\SetMathAlphabet{\mathsfit}{bold}{\encodingdefault}{\sfdefault}{bx}{n}

